\documentclass[journal=jacsat,manuscript=article]{achemso}

\usepackage[version=3]{mhchem} 
\usepackage{amsmath}

\usepackage[dvipsnames]{xcolor}

\author{Zhonglin Cao}
\affiliation[biogen]
{Medicinal Chemistry, Biogen, Cambridge, Massachusetts 02142, United States}

\author{Simone Sciabola}
\affiliation[biogen]
{Medicinal Chemistry, Biogen, Cambridge, Massachusetts 02142, United States}

\author{Ye Wang}
\affiliation[biogen]
{Medicinal Chemistry, Biogen, Cambridge, Massachusetts 02142, United States}

\email{ye.wang@biogen.com}
\title[]{Large-scale Pretraining Improves Sample Efficiency of Active Learning based Molecule Virtual Screening}

\abbreviations{IR,NMR,UV}

\begin{document}



\begin{abstract}
Virtual screening of large compound libraries to identify potential hit candidates is one of the earliest steps in drug discovery. As the size of commercially available compound collections grows exponentially to the scale of billions, brute-force virtual screening using traditional tools such as docking becomes infeasible in terms of time and computational resources. Active learning and Bayesian optimization has recently been proven as effective methods of narrowing down the search space. An essential component in those methods is a surrogate machine learning model that is trained with a small subset of the library to predict the desired properties of compounds. Accurate model can achieve high sample efficiency by finding the most promising compounds with only a fraction of the whole library being virtually screened. In this study, we examined the performance of pretrained transformer-based language model and graph neural network in Bayesian optimization active learning framework. The best pretrained models identifies 58.97\% of the top-50000 by docking score after screening only 0.6\% of an ultra-large library containing 99.5 million compounds, improving 8\% over previous state-of-the-art baseline. Through extensive benchmarks, we show that the superior performance of pretrained models persists in both structure-based and ligand-based drug discovery. Such model can serve as a boost to the accuracy and sample efficiency of active learning based molecule virtual screening.
\end{abstract}

\section{Introduction}
As an important step in early-stage drug discovery, structure-based virtual screening methods, such as molecular docking\cite{li2019overview, irwin2016docking, yuriev2015improvements} and molecular dynamics simulation\cite{dror2011pathway, buch2011complete}, predict the conformation and pose of a ligand in target protein binding pockets and formulate a quantitative measurement of binding affinity\cite{li2019overview}. Among all structure-based virtual screening methods, molecular docking estimate the protein-ligand binding affinity by optimizing a parameterized scoring function\cite{friesner2004glide, trott2010autodock, ewing2001dock}. Molecular docking is one of the most popular choices thanks to its relatively lower computational cost\cite{irwin2016docking, Graff2022, gorgulla2020open} (usually a few CPU seconds for each ligand\cite{yang2021efficient}) and acceptable accuracy. Many successes are witnessed in identifying potential drug candidates\cite{richter2012diazepam, min2012structure, chen2009molecular, teotico2009docking} using molecular docking. 

In the past decade, the size of synthesizable on-demand compound libraries has grown exponentially. For example, the size of ZINC database has increased from 120 million molecules in 2015\cite{sterling2015zinc} to more than 1 billion molecules in 2020\cite{irwin2020zinc20}. Commercially available Enamine \textit{REAL} database\cite{enamine} now includes 6 billion synthetically feasible compound selected from a larger Enamine \textit{REAL} Space containing 36 billion compounds\cite{bellmann2022comparison}. As the the number of organic molecules with 30 heavy atoms is estimated to be $10^{60}$\cite{bohacek1996art}, the ultralarge compound libraries can be expected to expand in the future. Although the sheer size of those libraries allows the possibility of identifying new and non-proprietary drug candidates, it brings challenges to the virtual screening campaign. Exhaustive virtual screening using molecular docking on the ultralarge libraries can be a Herculean task or even infeasible due to required computational resources and time\cite{grebner2019virtual, gorgulla2021multi, lyu2019ultra, sadybekov2022synthon}. Therefore, docking strategies that can accelerate screening of ultralarge libraries or reduce the screening space without compromising hit recovery rate will be beneficial to the drug discovery industry. 

The rapid advances of machine learning provide researchers a powerful tool to accelerate docking-based virtual screening. Machine learning models are trained as a faster surrogate of molecular docking to either predict the docking score\cite{berenger2021lean} or classify if a compound is a hit\cite{gentile2020deep, gentile2022artificial}. In the application of machine learning to molecule virtual screening, the accuracy of surrogate model can directly influence the hit recovery rate. The data used for the training of machine learning model is obtained through docking a subset of the library. Larger training set generally leads to higher prediction accuracy but docking more compounds increases the computation burden. Optimizing the trade-off between computation cost and model accuracy is a problem within the scope of active learning. Active learning is a subfield of machine learning that aims to train accurate model using minimum amount of data by allowing the model to actively sample unlabelled data to be added into the training set\cite{settles.tr09, di2023active}. Many previous works have shown the applicability of active learning in sample efficient molecule virtual screening. Yang et al.\cite{yang2021efficient} demonstrate that active learning with graph convolutional neural network can recover $>90\%$ by docking only 5\% of the library. Deep Docking method proposed by Gentile et al.\cite{gentile2020deep, gentile2022artificial} enables the virtual screening on 1.36 billion compounds by 100-fold data reduction. Graff et al.\cite{graff2021accelerating} develop a framework (MolPAL) based on batched Bayesian optimization\cite{shahriari2015taking, pyzer2018bayesian}, which formulate a strong synergy with pool-based active learning\cite{zhan2021comparative, di2023active}, to successfully recover 94.8\% of the top-50000 compounds from a 99.5 million sized library. Such a framework is shown to be effective in noisy environment\cite{bellamy2022batched} (a common problem with docking data), and a pruning algorithm is developed to further improve efficiency by reducing the screening space\cite{graff2022self}.

The choice of surrogate machine learning model in the active learning framework is one of the dominant factors that affect the hit recovery rate. A more accurate surrogate model can achieve high sample efficiency by allowing the active learning virtual screening to identify more hit candidates with less docking necessary. Self-supervised pretraining is a well-established technique to improve the models' performance in downstream tasks like regression. The goal of self-supervised pretraining to enable model to learn a better representation of data using the large quantity of unlabeled data. With the growth of molecular data, molecular representation learning through self-supervision has attracted academic attention\cite{krishnan2022self}. Language models such as transformer pretrained using mask-language-modeling on large scale SMILES data\cite{chithrananda2020chemberta, ahmad2022chemberta, ross2022large} show promising accuracy in molecular property prediction. Graph neural network (GNN) can also be pretrained in a contrastive learning manner\cite{you2020graph} using molecular graph augmentation\cite{wang2022molecular, wang2022improving}. Moreover, joint pretraining across graph/text or 2D/3D data has been proven to be beneficial to models in molecular or material property prediction\cite{zhu2021dual, stark20223d, cao2023moformer}. The success of pretrained models in molecular representation learning inspires us to leverage them for more sample-efficient virtual screening. In this work, we benchmark two pretrained models including molecular language transformer\cite{ross2022large} (MoLFormer) and molecular contrastive learning pretrained graph isomorphism network\cite{wang2022molecular, xu2018powerful} (MolCLR) in the MolPAL framework for molecule virtual screening. We demonstrate that pretrained models can achieve consistently higher hit recovery rate and enrichment factor on 50 thousand to 99.5 million compound datasets compared with a strong baseline, directed message passing neural network\cite{yang2019analyzing} (D-MPNN). We also compare the models' performance with different acquisition functions, including greedy and upper confidence boundary (UCB), and study the effect of uncertainty level on hit recovery rate and chemical diversity of identified compounds. At last, we extend the applicability of MolPAL with pretrained surrogate models to ligand-based drug discovery by virtual screening large library based on the 3D similarity to known active compounds. Pretrained deep learning models are shown to be superior surrogate model choices in both structure-based and ligand-based active learning virtual screening. 

\section{Results and discussion}
\begin{figure}
    \centering
    \includegraphics[width=\linewidth]{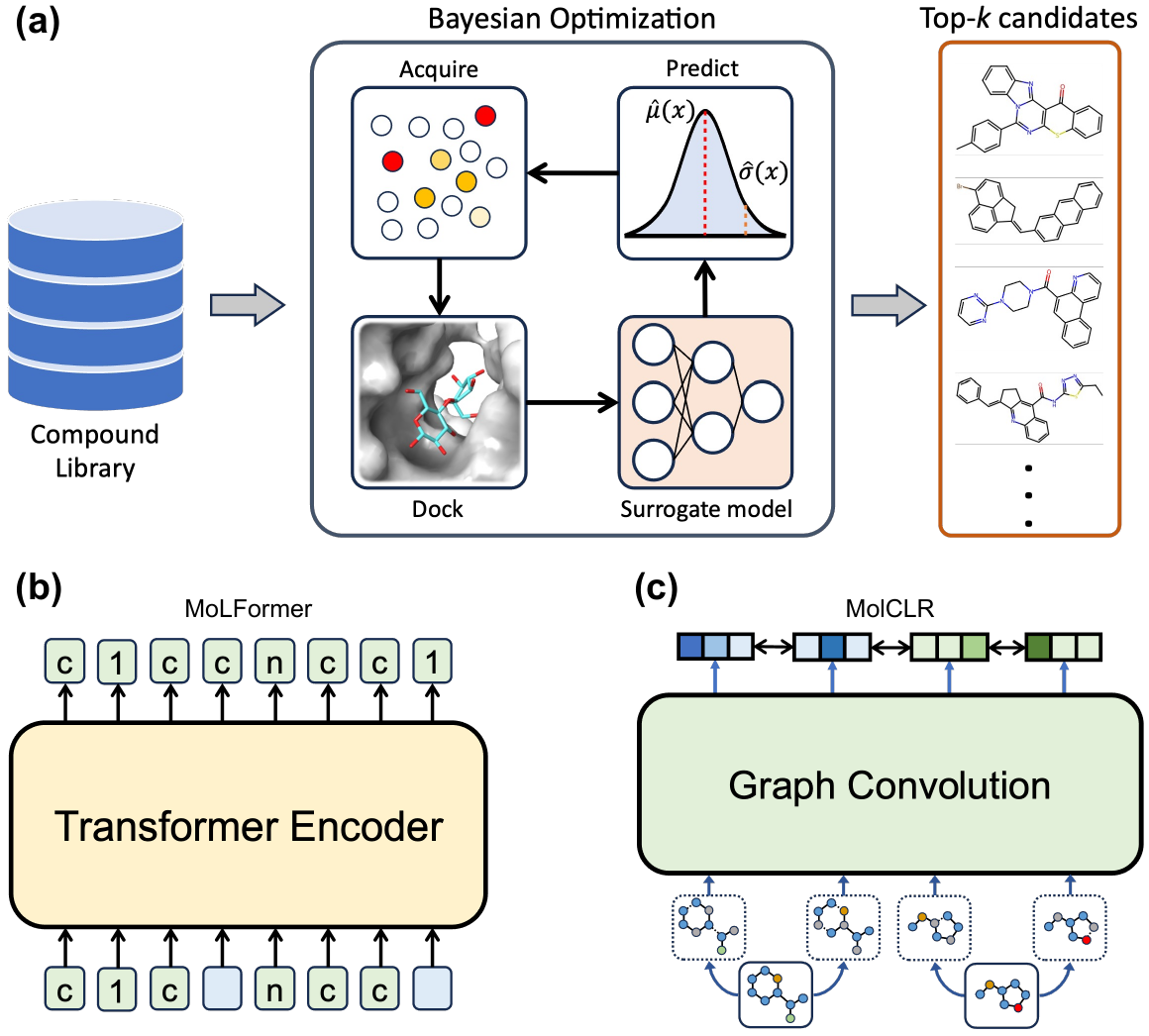}
    \caption{(a) The Bayesian optimization based active learning framework. For ligand-based virtual screening, the docking can be substitute by ligand similarity calculation such as Rapid Overlay of Chemical Structures (ROCS). (b) Schematic of mask-language-modeling pretraining process of the MoLFormer model. Green pads are tokenized SMILES, and blue pads represent masked tokens. MoLFormer is trained to predict masked tokens. (c) Schematic of the molecular contrastive learning pretraining process of the MolCLR model. Solid boxes contain original molecular graphs and dashed boxes contains augmented graphs. Grey circles represent masked nodes and dashed edges represent deleted bond during the augmentation process. Contrastive loss is applied on representations of augmented molecular graphs.}
    \label{fig:bo}
\end{figure}
The active learning is conducted using the batched Bayesian optimization framework, MolPAL, implemented by Graff et al\cite{graff2021accelerating} (Figure~\ref{fig:bo}a, details in the Method section). MolPAL consists of three important components including a surrogate model, acquisition function, and an objective function. The objective function can be a docking protocol in structure-based drug discovery. During the virtual screening of a large pool of molecules for potential hit to a protein target, a small batch of molecules are randomly selected and docked. The docking scores are used to train the surrogate model in a supervised manner such that the surrogate model can be used to predict the docking scores of all other molecules in the pool. The acquisition function evaluates the molecule pool based on the prediction from surrogate model and further selects another batch of molecules to augment the surrogate model training dataset. The above iterative process repeats until a user-defined stopping criterion, which is a fixed number of iterations in our case, is satisfied. We conduct retrospective studies on all datasets, meaning that the docking score of all molecules are pre-calculated and an oracle lookup function is used to retrieve the docking score during the active learning process instead of performing docking on-the-fly. Two acquisition functions tested in this work are the Greedy (Eq.\ref{eq:greedy}) and upper confidence boundary (UCB, Eq.~\ref{eq:ucb}). Details of all surrogate models and the training process are included in the Method section. Pretraining strategies of MoLFormer and MolCLR are visualized in Figure~\ref{fig:bo}b and \ref{fig:bo}c. The metrics we use to evaluate the performance of the Bayesian optimization active learning is the percentage of top-\textit{k} molecules retrieved (identified through SMILES) or top-\textit{k} retrieval rate and the enrichment factor (EF). The EF is defined as the percentage of top-\textit{k} molecules retrieved by active learning over the percentage of top-\textit{k} molecules retrieved by random selection.

\subsection{Enamine libraries}

\begin{figure}[htb!]
\includegraphics[width=0.7\linewidth]{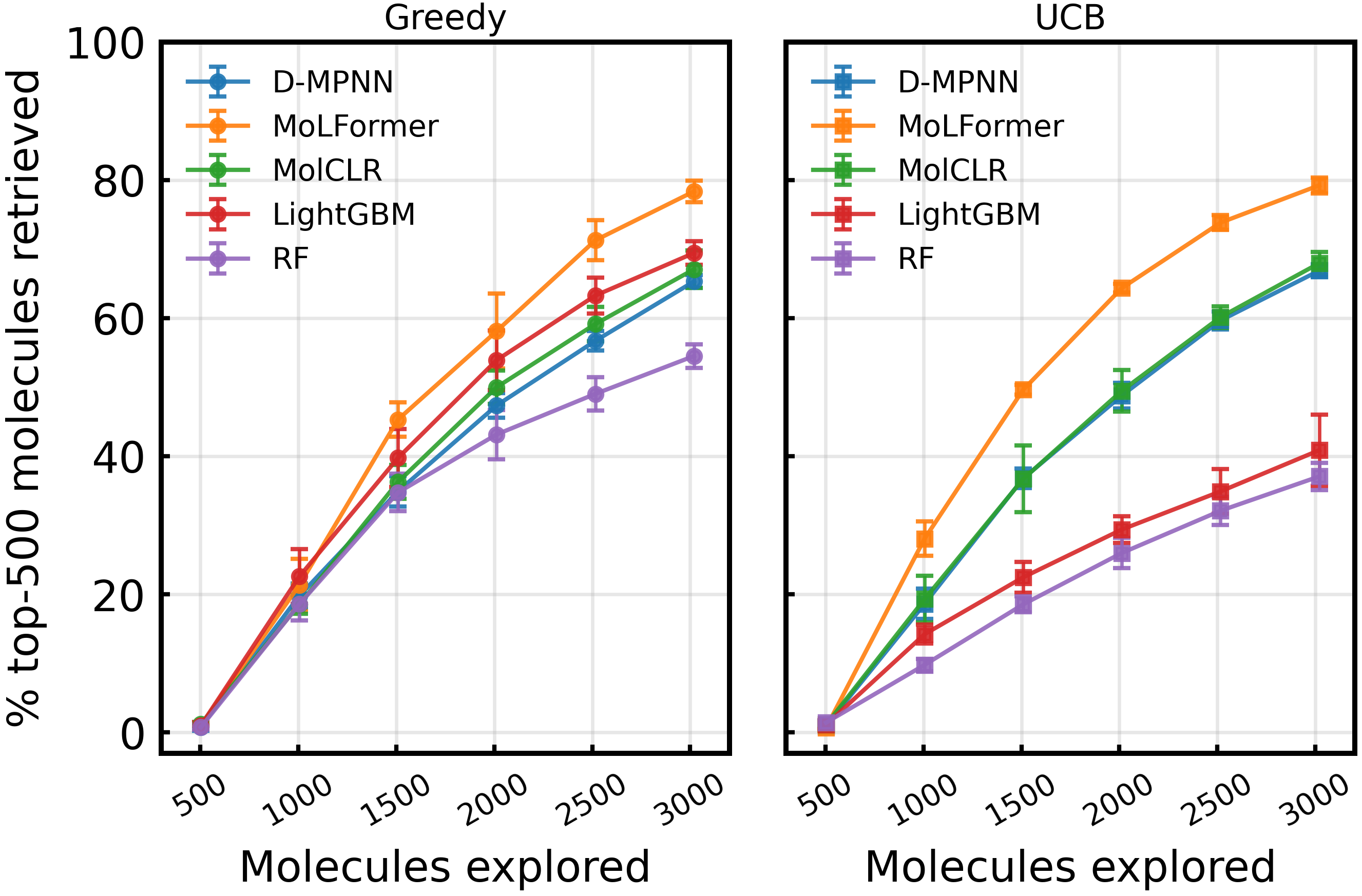}
\caption{Percentage of the top-500 (top-1\%) molecules in the Enamine 50K dataset retrieved after 5 iterations of Bayesian optimization using different surrogate models. The initial training set of the surrogate model contains 1\% of randomly selected molecules. Each following acquisition selects extra 1\% molecules according to the Greedy (left) or the UCB (right) strategy. Each data point is an average of 5 runs with different initial seeds and error bars represent one standard deviation.}
\label{fig:50k}
\end{figure}
Two Enamine compound libraries are used for benchmark in this work. The smaller one is the Enamine Discovery Diversity Set which contains 50,240 compounds (Enamine 50k). The larger one is the Enamine HTS collection (Enamine HTS) containing 2,141,514 molecules. The Enamine 50K and HTS datasets are publicly available on the code repository of MolPAL\cite{graff2021accelerating}. Molecules in both of the Enamine datasets are docked using AutoDock Vina\cite{trott2010autodock} against thymidylate kinase (PDB ID: 4UNN)\cite{naik2015structure}. The docking procedure is detailed in ref\cite{graff2021accelerating}. When the models are evaluated on the Enamine 50k, 1\% of the dataset is randomly selected for the initial model training. Following the initialization, 5 iterations of 1\% batch acquisition are done resulting in 6\% of the dataset being explored. MoLFormer retrieves 78.36\% of the top-500 molecules after 5 iterations of acquisition, followed by LightGBM (69.44\%), MolCLR (67.08\%), D-MPNN (65.32\%), and RF (54.52\%), using the Greedy strategy (Figure~\ref{fig:50k} left). UCB strategy slightly improves the top-500 retrieval rate of deep learning models, MoLFormer, MolCLR, and D-MPNN to 79.24\%, 67.96\%, and 66.88\%, respectively. However, the UCB negatively impacts LightGBM and RF and causes their top-500 retrieval rate drop to 40.88\% and 37.08\%, respectively (Figure~\ref{fig:50k} right). Calculated based on the higher retrieval rate of either Greedy or UCB acquisition strategy (Table S1), the EF of MoLFormer is 13.2, exceeding MolCLR (11.33), D-MPNN (11.15), LightGBM (11.57), and RF (9.09). The difference in the performance of GNN models, MolCLR and D-MPNN, is very small on the Enamine 50k dataset. LightGBM when used with the Greedy strategy is the second only to the MoLFormer. Considering its low computational cost, LightGBM is a promising surrogate model for screening small libraries.

\begin{figure}[htb!]
\includegraphics[width=\linewidth]{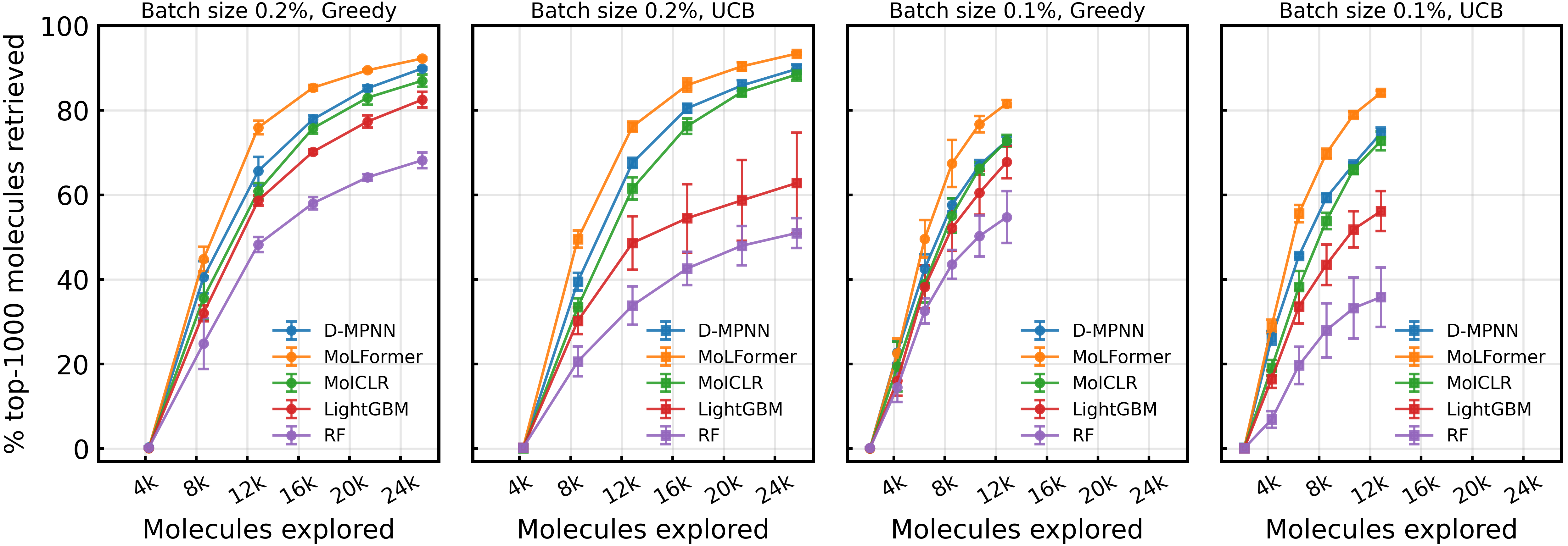}
\caption{Percentage of the top-1000 (top-0.05\%) molecules in the Enamine 50K dataset retrieved after 5 iterations of Bayesian optimization using different surrogate models. The initial training set of the surrogate model has the same size of the molecule batch acquired in each following iteration. The results of 0.2\% batch size (left two panels) and 0.1\% batch size (right two panels) are shown. Each data point is an average of 5 runs with different initial seeds and error bars represent one standard deviation.}
\label{fig:HTS}
\end{figure}

The same five surrogate models are further benchmarked on the larger Enamine HTS set (Figure~\ref{fig:HTS}) after their high EF is validated on the smaller Enamine 50k set. Given that the ultimate goal of active learning is to retrieve top performing compounds by minimum amount of evaluation, reduced batch size (0.2\% or 0.1\%) is used for the initial selection and iterative acquisition for the Enamine HTS set. Using the Greedy strategy and 0.2\% of acquisition batch size, MoLFormer retrieves 92.24\% of the top-1000 molecules by exploring only 1.2\% of the whole dataset, exceeding all other models. With the increase size of dataset, D-MPNN (89.88\%) and MolCLR (86.98\%) consistently outperform LightGBM (82.5\%) and RF (68.18\%). Deep learning models are generally better surrogate model choices when virtual screening large libraries. Smaller acquisition batch size improves the top-\textit{k} retrieval rate when the total number of explored compound remains the same. For example, after exploring 0.6\% of the dataset (12,855 molecules) using the Greedy strategy, MoLFormer retrieves 81.58\% of the top-1000 compounds with the 0.1\% batch size, higher than 75.9\% with the 0.2\% batch size. Smaller acquisition batch size also minimizes the performance gap between the D-MPNN and MolCLR. With Greedy strategy and 0.1\% batch size, MolCLR retrieves 72.78\% of the top-1000 compounds after exploring 0.6\% of the dataset, which is slightly higher than 72.76\% of the D-MPNN. When the batch size is 0.2\%, the top-1000 retrieval rate of D-MPNN (65.62\%) is noticeably higher than that of MolCLR (60.86\%) after exploring 0.6\% of the dataset. UCB strategy consistently improve the top-1000 retrieval rate of MoLFormer by 1-2.5\% regardless of the batch size on the Enamine HTS set. Similar to the results on the Enamine 50k set, top-1000 retrieval rates of LightGBM and RF drastically decrease by 11.6-19.7\% using the UCB instead of the Greedy strategy. However, neither Greedy nor UCB strategy is conclusively better than the other one based on the top-1000 retrieval rate comparison between D-MPNN and MolCLR (Table S2 and S3). Calculated based on the higher retrieval rate of either Greedy or UCB acquisition strategy with 0.1\% batch size, the EF of MoLFormer is 140.2, higher than D-MPNN (124.37), MolCLR (121.3), LightGBM (112.97), and RF (91.23).

\subsection{Effect of uncertainty weight in UCB acquisition}

\begin{figure}[htb!]
\includegraphics[width=0.9\linewidth]{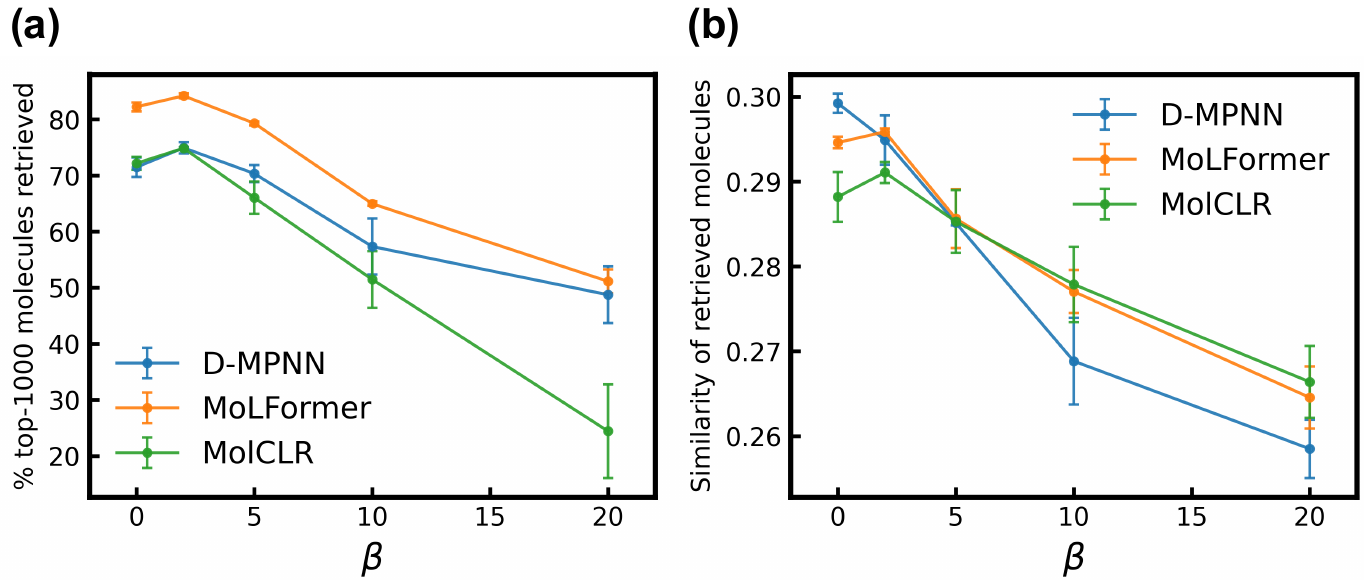}
\caption{The effect of increasing uncertainty weight ($\beta$) in the UCB acquisition function. (a) Top-1000 molecule retrieval rate of different surrogate models with increasing $\beta$ (b) Average Dice similarity of molecules retrieved by different surrogate models with increasing $\beta$. The results shown are from exploring 0.6\% of the EnamineHTS set with 0.1\% initialization and acquisition batch size. Each point is averaged value over three runs with different initial seed and error bar represents one standard deviation.}
\label{fig:beta}
\end{figure}

In the UCB acquisition function (Eq.~\ref{eq:ucb}), the weight of uncertainty in acquiring new samples is regulated by a hyperparameter $\beta$. To further understand the effect of $\beta$ on the top-\textit{k} molecule retrieval rate and the diversity of retrieved molecules, we run the BO active learning with $\beta$ value enumerating from ${0, 2, 5, 10, 20}$ on the Enamine HTS set. The results (Figure.~\ref{fig:beta}) are obtained using the 0.1\% batch size for initialization and iterative acquisition (0.6\% of total dataset explored at the end). When $\beta=0$, Greedy acquisition function is used. It is noticeable that the top-1000 retrieval rate peaks (Figure.~\ref{fig:beta}a) for all models when $\beta=2$, which is the default value of $\beta$, indicating that adding a term of uncertainty with small weight during acquisition can benefit the performance of the active learning. However, the top-1000 retrieval rate drops with greater $\beta$ value. For MoLFormer and D-MPNN, the retrieval rate converges to around 50\%, while the retrieval rate of MolCLR has negatively linear relationship with $\beta$ and drops to 24.4\% when $\beta=20$ without converging. To quantify the diversity of retrieved molecules, we compute the pairwise Dice similarity of molecules based on their Morgan Fingerprint with radius of 3. The average similarity between retrieved molecules drops with greater $\beta$ value, reflecting that the models tend to acquire more diverse molecules when uncertainty weight is higher in the acquisition function. If exploring a more diverse set of compound is desired during the active learning based virtual screening, tuning up the $\beta$ value is an effective option.

\subsection{Hundred-million scale libraries}
Industrial level virtual screening is usually conducted on ultra-large libraries containing hundred-millions or billions of compounds. To evaluate the performance of MolPAL with pretrained models on ultra-large library, we run the previous benchmarks on the dataset curated by Lyu et al.\cite{lyu2019ultra}. The dataset contains 99.5 million compounds docked against the AmpC $\mathrm{\beta}$-lactamase (the AmpC dataset in short, PDB ID: 1L2S) using Dock3.7\cite{ewing2001dock}. Since the deep learning surrogate models including MoLFormer, D-MPNN, and MolCLR significantly outperforms LightGBM and RF on large dataset like the Enamine HTS, only them are selected to be benchmarked on the AmpC dataset. The initialization and acquisition batch size is set to be 0.1\% of the whole dataset as we want to evaluate the models' performance given minimum amount of docking calculation. The ranking of top-50000 retrieval rate using the Greedy strategy is MolCLR (58.965\%) $>$ MoLFormer (55.497\%) $>$ D-MPNN (50.021\%). The UCB strategy reduces the top-50000 retrieval rate of MolCLR and MoLFormer to 55.659\% and 54.633\%, respectively, but improves that of the D-MPNN to 55.03\%. Among the three models (Table S4), MolCLR has the highest EF of 98.28, which is 6.2\% higher than MoLFormer (92.49) and 7.2\% higher than D-MPNN (91.72). By benchmarking on three datasets with various sizes, we demonstrate that pretrained deep learning surrogate models are more sample-efficient choices to be included in the Bayesian optimization active learning framework. 

\begin{figure}[htb!]
\includegraphics[width=0.7\linewidth]{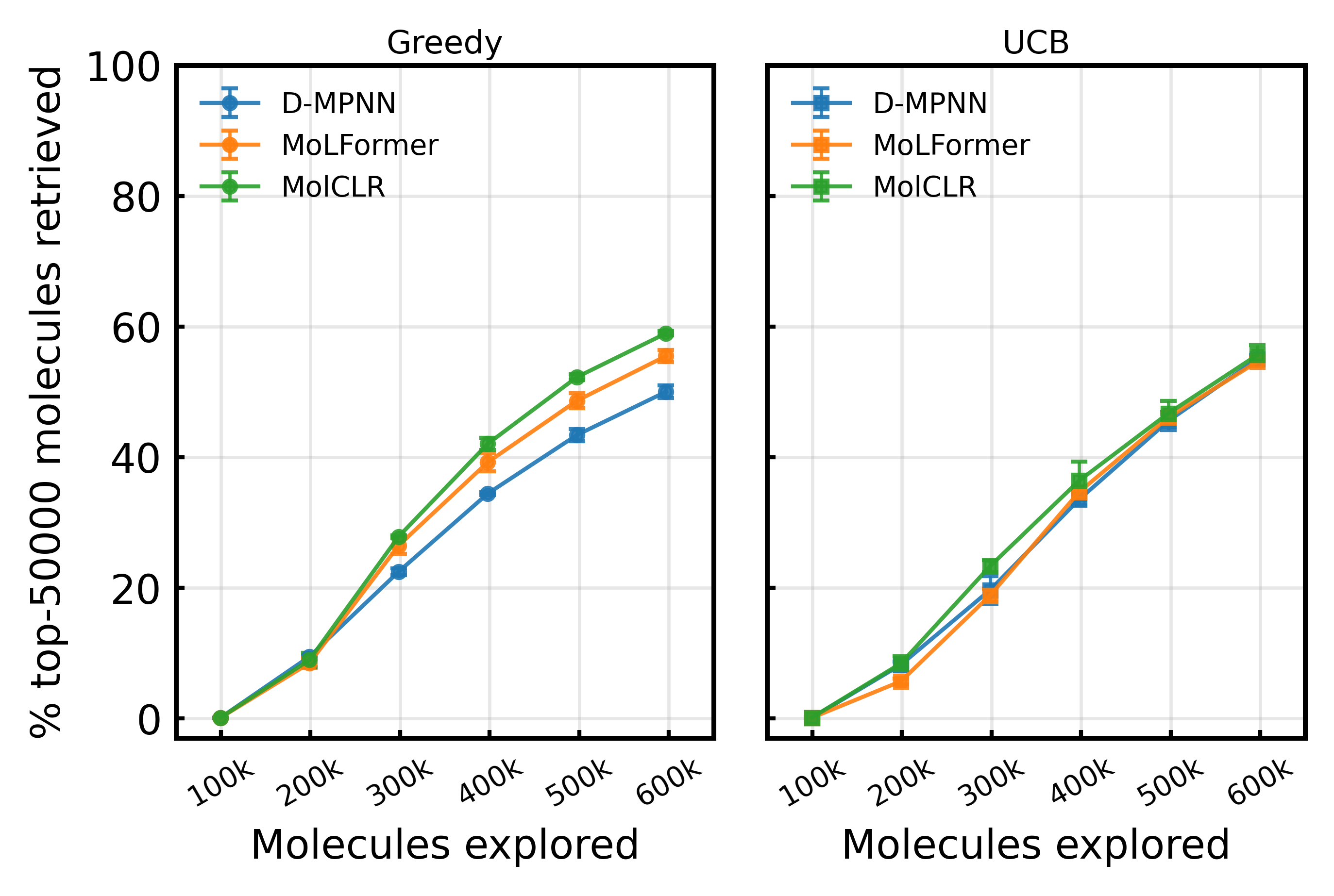}
\caption{Percentage of the top-50000 (top-0.05\%) molecules in the AmpC dataset retrieved after 5 iterations of Bayesian optimization using different surrogate models. The initial training set of the surrogate model contains 0.1\% of randomly selected molecules. Each following acquisition selects extra 0.1\% molecules according to the Greedy (left) or the UCB (right) strategy. Each data point is an average of 3 runs with different initial seeds and error bars represent one standard deviation.}
\label{fig:ampc}
\end{figure}

\subsection{Extending the application to ligand-based virtual screening}

\begin{figure}[htb!]
\includegraphics[width=0.7\linewidth]{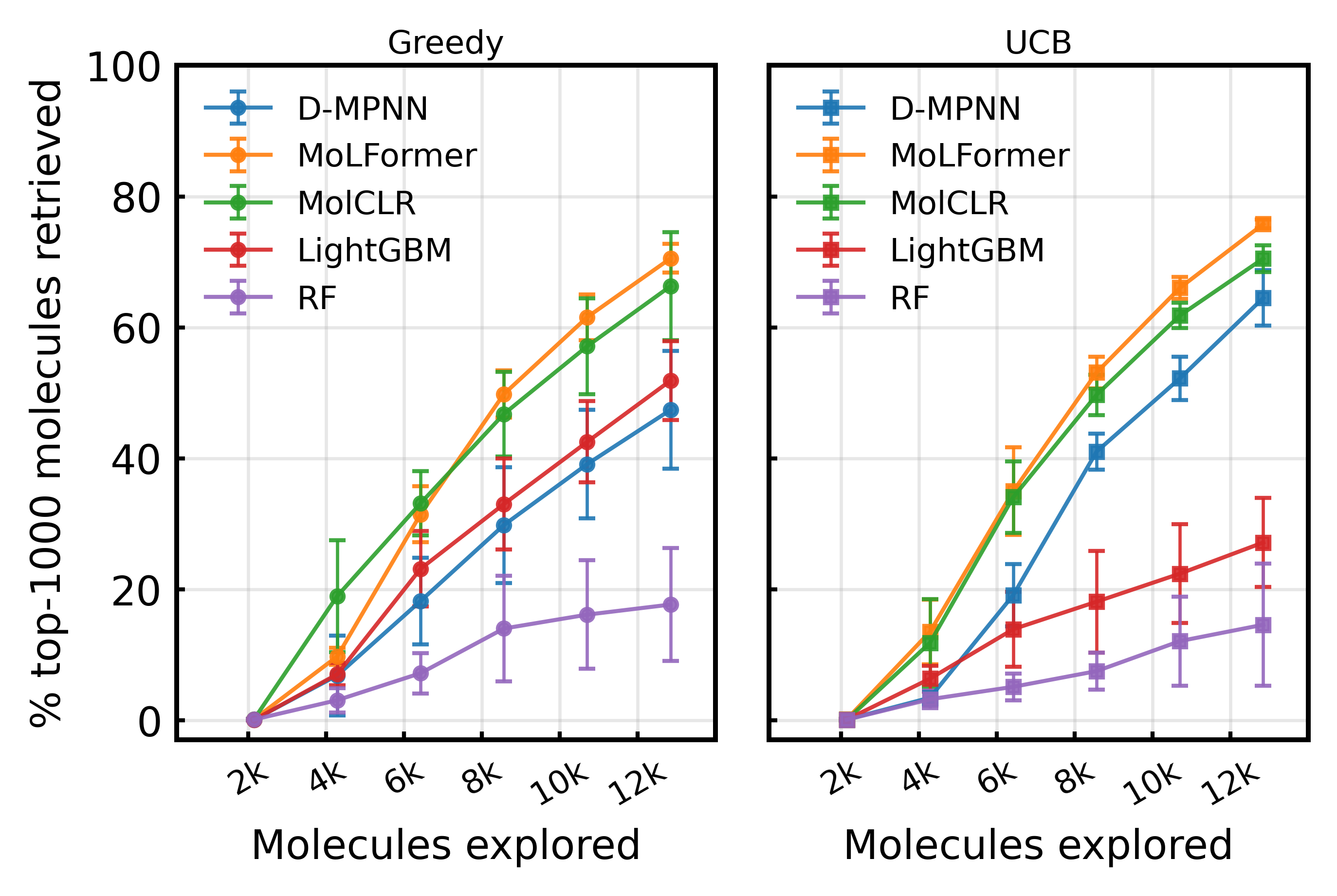}
\caption{Percentage of the top-1000 (top-0.05\%) molecules in the EnamineHTS ROCS dataset retrieved after 5 iterations of Bayesian optimization using different surrogate models. The initial training set of the surrogate model contains 0.1\% of randomly selected molecules. Each following acquisition selects extra 0.1\% molecules according to the Greedy (left) or the UCB (right) strategy. Each data point is an average of 5 runs with different initial seeds and error bars represent one standard deviation.}
\label{fig:HTSrocs}
\end{figure}

Ligand-based drug design is another important domain in the drug discovery industry. The fundamental philosophy of ligand-based drug design is that compounds similar to a known, active compound are more likely to be active. Rapid Overlay of Chemical
Structures (ROCS) by OpenEye Inc. is a established tool to calculate the 3D shape similarity of compounds. To extend the applicability of the MolPAL to ligand-based virtual screening, we compute the ROCS similarity score of all compounds in the Enamine HTS set to the original ligand in the thymidylate kinase complex\cite{naik2015structure} (see Method section for ROCS details and Figure S1 for EnamineHTS ROCS score distribution). MolPAL with different surrogate models is then benchmarked on the EnamineHTS ROCS dataset for the top-1000 molecules retrieval rate. The top-1000 molecules of the EnamineHTS ROCS dataset have ROCS score of at least 1.1771. 0.1\% batch size is used for the initialization and iterative acquisition. With the Greedy strategy (Figure~\ref{fig:HTSrocs} left), MoLFormer has the highest top-1000 retrieval rate (70.54\%), followed by MolCLR (66.30\%). The pretrained models have significantly higher top-1000 retrieval rate than LightGBM (51.88\%), D-MPNN (47.40\%), and RF (17.68\%). On the EnamineHTS ROCS dataset, the standard deviation of top-1000 retrieval rate across 5 runs is very large for all models except MoLFormer when Greedy strategy is used. Specifically, the retrieval rate standard deviation at the 5th iteration is higher than 8\% for both MolCLR and D-MPNN, indicating unstable performance. The UCB strategy effectively alleviates this problem for deep learning models (Figure~\ref{fig:HTSrocs} right). With UCB strategy, the average top-1000 retrieval rate of MoLFormer, MolCLR, and D-MPNN is improved to 75.78\%, 70.52\%, and 64.5\%, respectively. The retrieval rate standard deviation after the last iteration is also reduced to 0.66\% (MoLFormer), 2.04\% (MolCLR), and 4.24\% (D-MPNN) when following the UCB strategy. Given that UCB strategy can lead to higher and more consistent top-1000 retrieval rate for the deep learning models, it is the better choice when MolPAL is applied on ligand-based virtual screening with ROCS score.


\section{Conclusion}
In this work, we demonstrate that Transformer-based language models (MoLFormer) and graph neural network (MolCLR) after large-scale pretraining can serve as more sample-efficient surrogate model in the Bayesian optimization active learning framework for molecule virtual screening. On Enamine libraries with size ranging from 50 thousand to 2.1 million compounds, MoLFormer has consistently higher top-$k$ docking score molecules retrieval rate than the previous state-of-the-art D-MPNN model. On the AmpC dataset containing 99.5 million compounds docked against AmpC $\mathrm{\beta}$-lactamase, MoLFormer and MolCLR retrieves 5.5\% and 8.9\% more top-50000 compounds than D-MPNN, respectively, after exploring the same 0.6\% of the whole dataset using Greedy strategy. We demonstrate that smaller acquisition batch size can improve the top-$k$ molecules retrieval rate. Greedy acquisition strategy is effective in most cases, rendering it a suitable default option. The effect of UCB strategy varies on different surrogate models and datasets. The larger uncertainty weight in the UCB acquisition function is shown to increase the diversity of the acquired molecules at the cost of reduced top-$k$ retrieval rate. At last, the application of MolPAL is extended to ligand-based virtual screening. We curate a dataset containing the compounds in the EnamineHTS dataset and their ROCS score to the ligand in the 4UNN complex. With only 0.6\% of the EnamineHTS ROCS dataset explored, MolFormer and MolCLR again outperforms D-MPNN by large margins in retrieving the top-$1000$ similar molecules. UCB strategy is shown to be a better option on the ligand-based virtual screening because it delivers higher and more consistent top-$k$ retrieval rate. In conclusion, deep learning models after pretraining on large-scale chemical space, such as MoLFormer, should be used as surrogate models in active learning molecule virtual screening because of their higher sample-efficiency. The application of pretrained models can benefit both the structure-based and ligand-based domains in the field of drug discovery.

\section{Method}
\subsection{Bayesian optimization}
The active learning is conducted using the bacthed Bayesian optimization framework implemented by Graff et al\cite{graff2021accelerating} called MolPAL. MolPAL consists of three important components including a surrogate model, acquisition function, and an objective function (e.g. docking or ROCS). When applied on virtual screening of a large library of molecules (denoted as a set $\mathcal{X}$), the goal of Bayesian optimization is to select a subset $\mathbf{x}$ that maximizes the objective $f(\mathbf{x})$, which can be the docking score or ROCS score. Due to the relatively high computational cost of objective function, a machine learning model parameterized by $\theta$, $f_{\theta}(\cdot)$, is used as a faster surrogate to approximate the original objective function. 

At the beginning of active learning, a batch of molecules $\mathbf{x}_0$ are randomly selected and their corresponding objective values $f(\mathbf{x}_0)$ are calculated. The surrogate model is trained using the initial labeled training set $\mathcal{D}_0 = \{\mathbf{x}_0, f(\mathbf{x}_0)\}$. Then, for $n$ iterations, new batches of molecules, $\mathbf{x}_i$, will be selected by $\mathbf{x}_i = \underset{\mathbf{x}\in \mathcal{X}}{\mathrm{argmax}} \; \alpha(\mathbf{x}, f_{\theta}(\mathbf{x}))$, where $0<i\leq n$ is the $i$-th iteration and $\alpha(\cdot)$ is the acquisition function. After each iteration of selection, the training set will be augmented as $\mathcal{D}_i = \{ \mathcal{D}_{i-1}, (\mathbf{x}_i, f(\mathbf{x}_i))\}$ and the surrogate model will be retrained using $\mathcal{D}_i$. In this work, $n=5$ is set for all benchmarks. The top-$k$ molecules retrieval rate is calculated as the percentage of real top-$k$ molecules being included in the $\mathcal{D}_n$. Retrospective studies are conducted on all datasets, meaning that docking score of all molecules are pre-calculated and $f(\mathrm{x})$ is a oracle lookup function that retrieves the docking or ROCS score during the active learning process. 

\subsection{Acquisition functions}
Based on the benchmarks in the work of Graff et al.\cite{graff2021accelerating}, we test two of the best performing acquisition functions including the greedy strategy:
\begin{equation}
    \alpha_{\mathrm{greedy}}(x) = \hat{\mu}(x)
\label{eq:greedy}
\end{equation}
and the upper confidence boundary (UCB) strategy\cite{auer2002using,srinivas2009gaussian}:
\begin{equation}
    \alpha_{\mathrm{UCB}}(x) = \hat{\mu}(x) + \beta\hat{\sigma}(x)
\label{eq:ucb}
\end{equation}
where $\hat{\mu}(x)$ and $\hat{\sigma}(x)$ are the predicted docking or ROCS score and predicted uncertainty of a molecule $x$, respectively. Since new batches are selected by maximizing the acquisition function, docking scores are multiplied by -1. $\beta$, set toa default value of 2 in our work, is a hyperparameter that determines the weight of uncertainty when augmenting the surrogate model training data.

\subsection{Surrogate models}
Five surrogate models benchmarked in this work are Molecular Language Transformer\cite{ross2022large} (MoLFormer), directed message passing neural
network\cite{yang2019analyzing} (D-MPNN), graph isomorphism network\cite{xu2018powerful} pretrained using molecular contrastive learning\cite{wang2022molecular} (MolCLR), LightGBM\cite{ke2017lightgbm}, and Random Forest\cite{breiman2001random} (RF). Deep learning models, including MoLFormer, D-MPNN, and MolCLR, consist of two components, a feature extractor and a prediction head. The feature extractor (Transformer encoder in MoLFormer and graph neural network in D-MPNN and MolCLR) learns a molecular representation from the input SMILES. The prediction head is a two-layer fully connected neural network that predicts the docking score and uncertainty using the learned molecular representation. For LightGBM and RF, each molecule is represented by a 2048-bit atom-pair fingerprint\cite{carhart1985atom} with minimum radius of 1 and maximum radius of 3. When greedy acquisition function is used, the all surrogate models predict only the docking score of each molecule and are trained with the mean squared error (MSE) loss. When the UCB acquisition function is used, neural network based surrogate models predict both the docking score and the uncertainty (variance), and is trained on the gaussian negative log-likelihood loss\cite{nix1994estimating, hirschfeld2020uncertainty}:
\begin{equation}
    \mathcal{L}(y, \hat{y}, \hat{\sigma}^2) = \frac{1}{2}(\log \hat{\sigma}^2 + \frac{(\hat{y} - y)^2}{\hat{\sigma}^2})
\end{equation}
where the $y$ and $\hat{y}$ are the ground truth and predicted docking or ROCS score, respectively. $\hat{\sigma}^2$ is the predicted variance which is clamped to $10^{-5}$ for training stability. For random forest based models (RF and LightGBM), the uncertainty is calculated as the variance of predictions of all decision trees in the ensemble and they are always trained using the MSE loss function.

MoLFormer\cite{ross2022large} is a transformer-based\cite{vaswani2017attention} model. It receives tokenized\cite{schwaller2019molecular} SMILES string as input and generates molecular representation by learning the intrinsic spatial relationships between atoms in a molecule. Linear attention mechanism\cite{katharopoulos2020transformers} and rotary position embedding\cite{su2021roformer} are adopted over the regular quadratic attention to improve the computation efficiency of the model. MoLFormer has been pretrained on a ultra-large 1.1 billion small molecule dataset combining ZINC\cite{irwin2005zinc} and PubChem\cite{kim2019pubchem} using the mask-language-modeling technique\cite{devlin2018bert, liu2019roberta} (Figure~\ref{fig:bo}b). In this work, we use the MoLFormer-XL variant proposed in ref\cite{ross2022large} which consists of 12 linear attention layer with 12 heads in each layer and a hidden dimension of 768. Molecular representation learned by MoLFormer is the average of the embeddings of all input tokens. The prediction head of MoLFormer has hidden dimension as 768 with gaussian error linear unit\cite{hendrycks2016gaussian} activation and 0.1 rate of dropout\cite{srivastava2014dropout}. LAMB\cite{you2019large} optimizer is used to enable large batch training. The training batch size is 600 and learning rate is fixed at $1.6\times 10^{-4}$. 


D-MPNN and MolCLR pretrained graph isomorphism network\cite{xu2018powerful} are variants of the graph neural network\cite{scarselli2008graph} (GNN). They directly take the molecular graph, in which each atoms are represented as nodes and bonds as edges, as input thanks to their GNN architecture. Both model learn molecular representations through the message passing process. In message passing, the feature vector of each node in the molecular graph is updated by aggregating message from neighboring nodes. Through multiple aggregation update operations, the graph-level feature or molecular representation is calculated by a mean-pooling readout function over all node features in the graph. D-MPNN and MolCLR differs from each other by their unique aggregation update operation, which are detailed in ref\cite{yang2019analyzing} and ref\cite{xu2018powerful, wang2022molecular}, respectively. MolCLR is pretrained on approximately 10 million molecules from PubChem\cite{kim2019pubchem} using contrastive learning strategy\cite{chen2020simple} (Figure~\ref{fig:bo}c). For fair comparison, we use the default architecture setting of both the D-MPNN (3 graph layers and hidden size as 300) and MolCLR (5 graph layers, embedding size as 300, and feature size as 512). The prediction head of D-MPNN has hidden size of 300 and ReLU activation function. Following ref\cite{graff2021accelerating}, D-MPNN is trained 50 epochs in batches of 50 using Adam\cite{kingma2014adam} optimizer with Noam learning rate scheduler\cite{vaswani2017attention} ($10^{-4}$, $10^{-3}$, and $10^{-4}$ as the initial, maximum, and final learning rate, respectively). The prediction head of MolCLR has hidden size of 256 and 128 for the first and second layer, respectively, and ReLU activation function. MolCLR is trained 50 epoches in batches of 32 using Adam\cite{kingma2014adam} optimizer. The learning rate is 0.0002 and 0.0005 for the pretrained graph neural network and prediction head, respectively. RDKit\cite{landrum2013rdkit} is used to convert SMILES to molecular graph. Molecules acquired at each iteration are split by 80\%-20\% to the training and validation set. Early stopping strategy with patient value of 10 on validation loss is adopted during the training of all neural network based models.

RF and LightGBM are decision tree based ensemble methods. RF is implemented using the scikit-learn\cite{scikit-learn} package. The \texttt{n\_estimator} value is set to 100, \texttt{max\_depth} set to 8, and \texttt{min\_samples\_leaf} set to 1. LightGBM is implemented using its dedicated python package\cite{ke2017lightgbm}. The \texttt{n\_estimator} value for LightGBM is set to 100, while the \texttt{max\_depth} is unlimited.


\begin{acknowledgement}

We thank Dibyendu Mondal, Adam Antoszewski, and Nupur Bansal for valuable discussion.

\end{acknowledgement}

\begin{suppinfo}

Top-$k$ retrieval rate at each active learning iteration for Enamine 50k, Enamine HTS and AmpC dataset. The distribution of Enamine HTS ROCS score. 

\end{suppinfo}

\newpage
\bibliography{ref}

\end{document}


\pagenumbering{arabic}
\renewcommand*{\thepage}{S\arabic{page}}
\maketitle
\tableofcontents
\newpage
\section{Top-$k$ retrieval rate at each active learning iteration}
\subsection{Top-500 retrieval rate of Enamine 50k dataset}

\begin{table}[htb!]
\renewcommand{\thetable}{S\arabic{table}}
  \centering
  \resizebox{\textwidth}{!}{\begin{tabular}{l|l|lllllll}
    \toprule
    Model & Acquisition & Iter 0 & Iter 1 & Iter 2 & Iter 3 & Iter 4 & Iter 5 \\
    \midrule
    D-MPNN & Greedy & 0.80 (0.51) & 19.80 (1.75) & 34.92 (2.20) & 47.36 (1.76) & 56.72 (1.43) & 65.32 (0.93)\\
    MoLFormer & Greedy & 1.00 (0.20) & 21.36 (3.76) & 45.32 (2.49) & 58.12 (5.43) & 71.28 (2.89) & 78.36 (1.58)\\
    MolCLR & Greedy & 1.24 (0.30) & 18.56 (1.37) & 36.28 (2.44) & 50.00 (2.44) & 59.16 (2.47) & 67.08 (2.70)\\
    LightGBM & Greedy & 1.00 (0.49) & 22.60 (4.00) & 39.76 (4.20) & 53.92 (4.31) & 63.28 (2.58) & 69.44 (1.72)\\
    RF & Greedy & 0.76 (0.38) & 18.60 (2.36) & 34.76 (2.72) & 43.16 (3.58) & 49.04 (2.41) & 54.52 (1.72)\\
    \midrule
    D-MPNN & UCB & 0.80 (0.51) & 19.80 (1.75) & 34.92 (2.20) & 47.36 (1.76) & 56.72 (1.43) & 65.32 (0.93)\\
    MoLFormer & UCB & 1.00 (0.20) & 21.36 (3.76) & 45.32 (2.49) & 58.12 (5.43) & 71.28 (2.89) & 78.36 (1.58)\\
    MolCLR & UCB & 1.24 (0.30) & 18.56 (1.37) & 36.28 (2.44) & 50.00 (2.44) & 59.16 (2.47) & 67.08 (2.70)\\
    LightGBM & UCB & 1.00 (0.49) & 22.60 (4.00) & 39.76 (4.20) & 53.92 (4.31) & 63.28 (2.58) & 69.44 (1.72)\\
    RF & UCB & 0.76 (0.38) & 18.60 (2.36) & 34.76 (2.72) & 43.16 (3.58) & 49.04 (2.41) & 54.52 (1.72)\\
    \bottomrule
  \end{tabular}}
  \caption{Top-500 retrieval rate of Enamine 50k dataset from initialization to the 5th iteration of active learning. Values reported are the average over 5 runs. Numbers in the parenthesis are one standard deviation. 1\% batch size is used for both initialization and iterative acquisition.}
  \label{tb:50k}
\end{table}
\newpage

\subsection{Top-1000 retrieval rate of Enamine HTS dataset}

\begin{table}[htb!]
\renewcommand{\thetable}{S\arabic{table}}
  \centering
  \resizebox{\textwidth}{!}{\begin{tabular}{l|l|lllllll}
    \toprule
    Model & Acquisition & Iter 0 & Iter 1 & Iter 2 & Iter 3 & Iter 4 & Iter 5 \\
    \midrule
    D-MPNN & Greedy & 0.30 (0.19) & 40.52 (3.82) & 65.62 (3.38) & 77.84 (0.93) & 85.20 (0.68) & 89.88 (0.31)\\
    MoLFormer & Greedy & 0.10 (0.00) & 44.80 (2.91) & 75.90 (1.60) & 85.34 (0.69) & 89.46 (0.25) & 92.24 (0.28)\\
    MolCLR & Greedy & 0.18 (0.04) & 35.64 (5.12) & 60.86 (1.94) & 75.74 (1.24) & 82.98 (1.64) & 86.98 (1.44)\\
    LightGBM & Greedy & 0.20 (0.10) & 32.04 (1.90) & 58.66 (1.18) & 70.18 (0.56) & 77.34 (1.44) & 82.50 (1.88)\\
    RF & Greedy & 0.26 (0.21) & 24.84 (6.02) & 48.28 (1.79) & 58.02 (1.47) & 64.16 (0.73) & 68.18 (1.85)\\
    \midrule
    D-MPNN & UCB & 0.16 (0.13) & 39.48 (2.07) & 67.58 (1.17) & 80.46 (1.09) & 85.86 (1.26) & 89.74 (1.13)\\
    MoLFormer & UCB & 0.18 (0.04) & 49.54 (2.06) & 76.14 (1.18) & 85.94 (1.52) & 90.38 (0.96) & 93.38 (0.33)\\
    MolCLR & UCB & 0.10 (0.12) & 33.44 (2.08) & 61.52 (2.61) & 76.24 (1.81) & 84.34 (1.08) & 88.44 (1.37)\\
    LightGBM & UCB & 0.22 (0.08) & 30.20 (3.16) & 48.64 (6.31) & 54.48 (8.05) & 58.70 (9.53) & 62.78 (11.96)\\
    RF & UCB & 0.24 (0.09) & 20.62 (3.54) & 33.84 (4.52) & 42.60 (3.91) & 47.98 (4.62) & 50.96 (3.54)\\
    \bottomrule
  \end{tabular}}
  \caption{Top-1000 retrieval rate of Enamine HTS dataset from initialization to the 5th iteration of active learning. Values reported are the average over 5 runs. Numbers in the parenthesis are one standard deviation. 0.2\% batch size is used for both initialization and iterative acquisition.}
  \label{tb:HTS_0.2}
\end{table}

\begin{table}[htb!]
\renewcommand{\thetable}{S\arabic{table}}
  \centering
  \resizebox{\textwidth}{!}{\begin{tabular}{l|l|lllllll}
    \toprule
    Model & Acquisition & Iter 0 & Iter 1 & Iter 2 & Iter 3 & Iter 4 & Iter 5 \\
    \midrule
    D-MPNN & Greedy & 0.10 (0.07) & 22.78 (2.55) & 42.54 (3.44) & 57.62 (1.55) & 66.98 (1.23) & 72.76 (1.07)\\
    MoLFormer & Greedy & 0.04 (0.05) & 22.52 (3.49) & 49.62 (4.43) & 67.42 (5.55) & 76.70 (1.93) & 81.58 (0.84)\\
    MolCLR & Greedy & 0.08 (0.04) & 19.40 (5.86) & 38.96 (4.38) & 55.12 (4.08) & 66.24 (1.39) & 72.78 (1.39)\\
    LightGBM & Greedy & 0.10 (0.10) & 16.08 (3.60) & 38.20 (5.06) & 52.18 (5.43) & 60.48 (5.19) & 67.78 (3.82)\\
    RF & Greedy & 0.14 (0.09) & 14.52 (3.50) & 32.58 (2.97) & 43.60 (3.42) & 50.24 (4.84) & 54.74 (6.14)\\
    \midrule
    D-MPNN & UCB & 0.22 (0.04) & 26.20 (1.60) & 45.60 (0.39) & 59.32 (1.03) & 67.20 (0.52) & 74.62 (1.31)\\
    MoLFormer & UCB & 0.12 (0.04) & 28.90 (1.61) & 55.58 (2.03) & 69.76 (1.13) & 78.98 (0.74) & 84.12 (0.58)\\
    MolCLR & UCB & 0.18 (0.13) & 19.10 (1.86) & 38.22 (3.79) & 53.84 (1.94) & 66.02 (1.11) & 72.76 (2.21)\\
    LightGBM & UCB & 0.06 (0.05) & 16.46 (2.09) & 33.58 (3.97) & 43.46 (4.77) & 51.84 (4.28) & 56.14 (4.72)\\
    RF & UCB & 0.08 (0.08) & 6.90 (1.98) & 19.70 (4.43) & 27.96 (6.42) & 33.24 (7.23) & 35.82 (7.04)\\
    \bottomrule
  \end{tabular}}
  \caption{Top-1000 retrieval rate of Enamine HTS dataset from initialization to the 5th iteration of active learning. Values reported are the average over 5 runs. Numbers in the parenthesis are one standard deviation. 0.1\% batch size is used for both initialization and iterative acquisition.}
  \label{tb:HTS_0.1}
\end{table}
\newpage

\subsection{Top-50000 retrieval rate of AmpC dataset}
\begin{table}[htb!]
\renewcommand{\thetable}{S\arabic{table}}
  \centering
  \resizebox{\textwidth}{!}{\begin{tabular}{l|l|lllllll}
    \toprule
    Model & Acquisition & Iter 0 & Iter 1 & Iter 2 & Iter 3 & Iter 4 & Iter 5 \\
    \midrule
    D-MPNN & Greedy & 0.10 (0.01) & 9.47 (0.30) & 22.47 (0.52) & 34.43 (0.22) & 43.39 (0.91) & 50.02 (0.94)\\
    MoLFormer & Greedy & 0.10 (0.01) & 8.48 (0.75) & 26.48 (1.24) & 39.23 (1.39) & 48.65 (1.17) & 55.50 (0.90)\\
    MolCLR & Greedy & 0.10 (0.02) & 8.97 (1.11) & 27.84 (0.17) & 42.05 (0.91) & 52.26 (0.46) & 58.97 (0.34)\\
    \midrule
    D-MPNN & UCB & 0.10 (0.01) & 8.22 (0.49) & 19.66 (2.10) & 33.53 (0.88) & 45.54 (1.39) & 55.03 (0.56)\\
    MoLFormer & UCB & 0.10 (0.01) & 5.69 (0.44) & 18.83 (1.02) & 34.68 (0.30) & 46.09 (0.50) & 54.63 (0.49)\\
    MolCLR & UCB & 0.12 (0.01) & 8.43 (1.11) & 23.31 (0.90) & 36.41 (2.93) & 46.71 (1.87) & 55.66 (1.51)\\
    \bottomrule
  \end{tabular}}
  \caption{Top-50000 retrieval rate of AmpC dataset from initialization to the 5th iteration of active learning. Values reported are the average over 3 runs. Numbers in the parenthesis are one standard deviation. 0.1\% batch size is used for both initialization and iterative acquisition.}
  \label{tb:ampc}
\end{table}
\newpage

\section{Enamine HTS ROCS score distribution}
\begin{figure}[htb!]
\renewcommand{\thefigure}{S\arabic{figure}}
    \centering
    \includegraphics[width=0.7\linewidth]{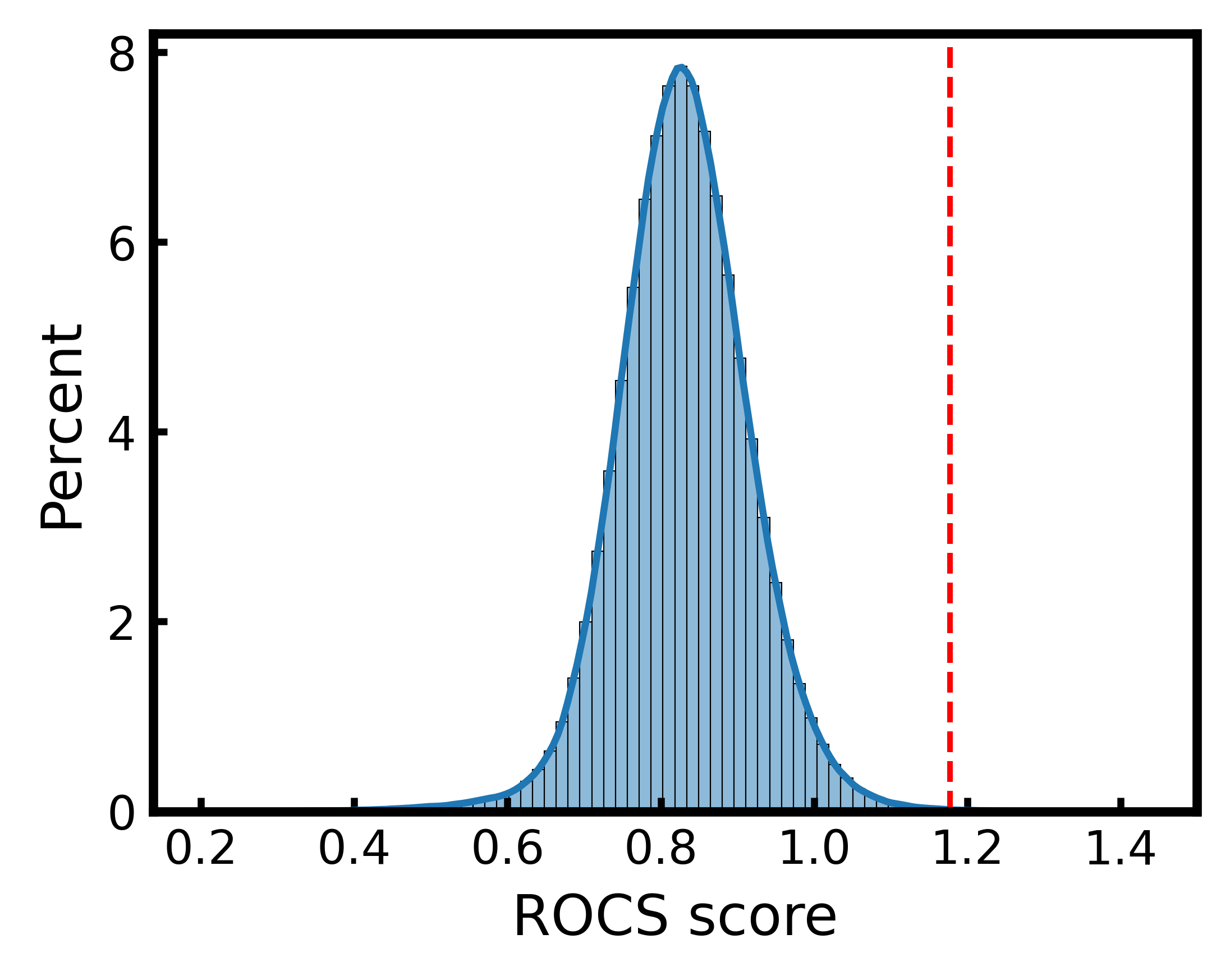}
    \caption{Distribution of the ROCS score of molecules in the Enamine HTS set. The cutoff score of the top-1000 molecules is 1.1771, marked by the red dashed line.}
    \label{fig:HTSrocs_dist}
\end{figure}


\newpage